\documentclass[letterpaper]{article} % DO NOT CHANGE THIS
\usepackage{aaai25}
\nocopyright
\usepackage{times}  % DO NOT CHANGE THIS
\usepackage{helvet}  % DO NOT CHANGE THIS
\usepackage{courier}  % DO NOT CHANGE THIS
\usepackage[hyphens]{url}  % DO NOT CHANGE THIS
\usepackage{graphicx} % DO NOT CHANGE THIS

\usepackage{tabularx} % For automatic column width adjustment
\usepackage{booktabs} % For better looking borders
\usepackage{colortbl} % For adding colors
\usepackage{array}    % For vertical centering and more control
\usepackage{xcolor}
\usepackage[utf8]{inputenc}
\usepackage{amsmath}
\usepackage{float}% For defining custom colors
\usepackage{svg}
\usepackage[T1]{fontenc}
\usepackage{tcolorbox} % for creating colored boxes
\usepackage{natbib}  % DO NOT CHANGE THIS AND DO NOT ADD ANY OPTIONS TO IT
\usepackage{caption} % DO NOT CHANGE THIS AND DO NOT ADD ANY OPTIONS TO IT
\usepackage{algorithm}
\usepackage{algorithmic}
\usepackage{comment}
\usepackage{xspace}
\newcommand{\modelname}{\textit{HALO}\xspace}

\usepackage{todonotes}
\usepackage{natbib}
\usepackage{enumitem}

\usepackage{newfloat}
\usepackage{listings}
\DeclareCaptionStyle{ruled}{labelfont=normalfont,labelsep=colon,strut=off} % DO NOT CHANGE THIS
\floatstyle{ruled}
\newfloat{listing}{tb}{lst}{}
\floatname{listing}{Listing}
\title{\modelname: Hallucination Analysis and Learning Optimization to Empower LLMs with Retrieval-Augmented Context for Guided Clinical Decision Making}
\author {
    Sumera Anjum\textsuperscript{\rm 1},
    Hanzhi Zhang\textsuperscript{\rm 1},
    Wenjun Zhou\textsuperscript{\rm 2},
    Eun Jin Paek\textsuperscript{\rm 3},
    Xiaopeng Zhao\textsuperscript{\rm 2},
    Yunhe Feng\textsuperscript{\rm 1}
}

\affiliations{
    \textsuperscript{\rm 1}University of North Texas, Denton, TX 76207, USA\\
    \textsuperscript{\rm 2}University of Tennessee, Knoxville, TN 37996, USA\\
    \textsuperscript{\rm 3}University of Tennessee Health Science Center, Knoxville, TN 37996, USA\\
    sumeraanjumsumeraanjum@my.unt.edu, hanzhizhang@my.unt.edu, wzhou4@utk.edu, epaek@uthsc.edu, xzhao9@utk.edu, yunhe.feng@unt.edu
}

\usepackage{bibentry}
\begin{document}

% Define custom color for table rows
\definecolor{lightgray}{gray}{0.95}

\maketitle

\begin{abstract}
Large language models (LLMs) have significantly advanced natural language processing tasks, yet they are susceptible to generating inaccurate or unreliable responses, a phenomenon known as hallucination. In critical domains such as health and medicine, these hallucinations can pose serious risks. This paper introduces \modelname, a novel framework designed to enhance the accuracy and reliability of medical question-answering (QA) systems by focusing on the detection and mitigation of hallucinations. Our approach generates multiple variations of a given query using LLMs and retrieves relevant information from external open knowledge bases to enrich the context. We utilize maximum marginal relevance scoring to prioritize the retrieved context, which is then provided to LLMs for answer generation, thereby reducing the risk of hallucinations. The integration of LangChain further streamlines this process, resulting in a notable and robust increase in the accuracy of both open-source and commercial LLMs, such as Llama-3.1 (from 44\% to 65\%) and ChatGPT (from 56\% to 70\%). This framework underscores the critical importance of addressing hallucinations in medical QA systems, ultimately improving clinical decision-making and patient care.
% \href{https://www.example.com}{Click here to visit Example}
The open-source \modelname is available at: \textcolor{blue}{\url{https://github.com/ResponsibleAILab/HALO}}. 
\end{abstract}

\section{Introduction}
In recent years, Large language models (LLMs) have emerged as powerful tools with remarkable capabilities across diverse applications, ranging from natural language understanding to text generation. However, despite their widespread application, LLMs often suffer from a critical issue known as hallucination, where they generate incorrect or unreliable answers. Several factors lead to these hallucinations. The training data for LLMs often includes misconceptions~\cite{lin2021truthfulqa}, social biases~\cite{ladhak2023pre}, duplication biases~\cite{lee2021deduplicating}, knowledge cutoffs~\cite{onoe2022entity}, and a lack of specific domain training~\cite{li2023chatdoctor}, all of which can lead to the model generating incorrect outputs. During training, the model is often pushed to provide answers beyond its knowledge, known as capability misalignment~\cite{ranzato2015sequence}. Belief misalignment, or sycophancy~\cite{burns2022discovering}, occurs when the model generates answers it believes the user wants to hear, regardless of their correctness, due to training on datasets that emphasize user satisfaction over factual accuracy. The prompts given to the LLM are also very crucial in generating hallucinations. Poorly written prompts~\cite{yu2022legal} fail to fully grasp the context of a query, leading to insufficient context attention~\cite{shi2023context}. It focuses too much on nearby words, resulting in inaccurate answers. In fields such as health and medicine, where precision and accuracy are paramount~\cite{umapathi2023med}, the use of hallucinated answers can be extremely dangerous because they directly impact patient safety, treatment effectiveness, and resource utilization~\cite{rosol2023evaluation}.

Existing solutions to mitigate hallucinations in LLMs have several limitations. For instance, Med-HALT~\cite{umapathi2023med} proposed a set of instructions to help identify hallucinations and prompt LLMs to avoid incorrect answers for reasoning- and memory-based medical questions. However, it did not explore strategies such as integrating external knowledge to enhance the model’s performance when it lacks the specific information needed to answer a question. In the financial domain, \citet{kang2023deficiency} evaluated LLM hallucination mitigation techniques, including few-shot learning. However, these methods were not tested across a wide range of real-world scenarios and demonstrated varying levels of effectiveness.
The Hypotheses-to-Theories framework~\cite{zhu2023large} improved LLM reasoning but was constrained by the model's knowledge and input capacity. Additionally, approaches addressing ``hallucination snowballing''~\cite{zhang2023language} are limited by API constraints, hindering the large-scale exploration of mitigation strategies. These drawbacks underscore the need for more comprehensive and effective hallucination mitigation solutions.

This paper addresses the challenge of improving the accuracy and reliability of medical question answering (QA) systems by proposing \modelname, a novel framework that enhances LLMs' contextual reasoning abilities. Building upon the Retrieval Augmented Generation (RAG)~\cite{ji2023survey,sharma2024retrieval}, our framework integrates innovative techniques tailored specifically for the medical domain. It leverages open-source LangChain\footnote{https://github.com/langchain-ai/langchain} to generate multiple queries from the same question, expanding the range of keywords and covering wider aspects of the query. Additionally, we integrate more relevant and trustworthy medical knowledge from PubMed\footnote{https://pubmed.ncbi.nlm.nih.gov/} to generate context on the question. PubMed is used for medical knowledge because it provides access to a vast and reputable database of peer-reviewed medical research, ensuring that the information used is accurate, up-to-date, and based on the latest scientific evidence~\cite{umapathi2023med}. To create high-quality context, we employ a maximum marginal relevance~\cite{carbonell1998use} scoring mechanism to balance the trade-off between relevance and diversity of retrieved PubMed data. The top-scored context is then provided to LLMs, along with a few-shot prompting strategy to guide the model's understanding of how to answer the question step by step. Finally, we evaluate the effectiveness of our framework on the MEDMCQA dataset~\cite{pal2022medmcqa}, which contains 194k high-quality AIIMS \& NEET PG entrance exam multiple-choice questions covering 2.4k healthcare topics and 21 medical subjects. Our experimental results demonstrate that \modelname can boost the trustworthiness and reliability of medical question answering tasks. 

Aiming to enhance clinical decision-making and ensure the reliability of information in medical QA scenarios, the key contributions of this work are summarized as follows:
\begin{itemize}
    \item We propose and implement \modelname, a novel open-source framework designed to enhance the accuracy and reliability of medical QA systems by focusing on hallucination detection and mitigation.
    \item \modelname boosts LLMs' capability of providing more contextually grounded responses by incorporating external knowledge with RAG and refining the LLMs' understanding of the query with carefully designed prompt engineering, such as few-shot prompting, chain-of-thought reasoning, and appropriate prioritization. 
    \item Extensive experiments on 194k medical questions across 21 subjects with both commercial LLM (ChatGPT) and open-source LLMs (Llama-3.1 and Mistral) demonstrate the generalizability and adaptability of the proposed \modelname. Additionally, a case study on neurological disorders highlights \modelname's effectiveness in addressing specific disease-related queries.
\end{itemize}

\section{Related Work}
Hallucinations in LLMs have been identified as a significant concern across various domains, such as medicine~\cite{umapathi2023med, pal2022medmcqa, ponce2024large, rehman2023hallucination}, finance~\cite{kang2023deficiency}, education~\cite{sharma2024retrieval}, and law~\cite{yu2022legal}. The potential consequences of these hallucinations, particularly in critical domains like healthcare, underscore the importance of addressing this issue.

Prompt engineering and instruction techniques like few-shot learning~\cite{kang2023deficiency,yu2022legal} and chain-of-thought (CoT)~\cite{yu2022legal,braunschweiler2023evaluating} prompts can guide LLMs to improve task performance and reasoning accuracy. Few-shot learning uses a few examples to enhance understanding, while CoT prompts encourage step-by-step reasoning to reduce errors. However, these methods increase complexity and computational cost, and their success is highly dependent on carefully designed prompts. \citet{kang2023deficiency} showed limitations of these methods due to heavy reliance on the prompt quality. Additionally, \citet{yu2022legal} reported significant performance variability across test sets, and \citet{braunschweiler2023evaluating} found the effectiveness of prompts was inconsistent and responses were often verbose.

Recent efforts to integrate external knowledge sources into LLMs have shown promise in improving factual accuracy. Approaches such as retrieval-augmented generation (RAG) \cite{sharma2024retrieval,thorne2018fever} and prompt-based tool learning \cite{kang2023deficiency} leverage external knowledge to enhance the reliability of generated responses. While these methods have demonstrated success, they require carefully designed prompts and significant computational resources. Additionally, their effectiveness is often constrained by their domain-specific nature and reliance on up-to-date data sources, which may not comprehensively address general real-world scenarios.

Validation and explanation techniques like Verification-of-Choice (VoC)~\cite{li2023beginner, zhu2023large} and self-reflection~\cite{zhang2023language, zhu2023large} can enhance reasoning and self-correction in LLMs to mitigate hallucinations. VoC generates and verifies explanations for each choice, while self-reflection allows models to correct their errors. However, these methods are computationally expensive and their success is highly dependent on the initial accuracy of the model. Their effectiveness is constrained by the model's initial error rate and requires a robust base model, which is also limited by context length restrictions.

To improve the reliability and accuracy of LLMs in specific domains, various domain-specific prompt tuning and model fine-tuning techniques have been proposed. For instance, \citet{zhu2023promptcblue} evaluated fine-tuning strategies for Chinese biomedical tasks, while \citet{jeong2024olaph} fine-tuned LLMs using biomedical knowledge to improve factual accuracy. However, these approaches rely on pre-existing knowledge with fixed timestamps, limiting access to the most recent information and reducing their generalizability to evolving real-world scenarios.

While existing techniques for mitigating hallucinations in LLMs have made significant progress, each has notable limitations. Prompt engineering and instruction methods, such as few-shot learning and CoT prompts, are highly dependent on prompt quality and introduce additional complexity. External knowledge integration methods like RAG demand substantial computational resources, while validation techniques, including VoC and self-reflection, are resource-intensive and rely on the initial accuracy of the model. Domain-specific fine-tuning, although effective, often lacks generalizability and requires considerable resources. We propose \modelname to harness the strengths of these hallucination mitigation methods while reducing their limitations. \modelname improves LLM reliability performance by retrieving and prioritizing relevant information from domain-specific datasets, applying a maximum marginal relevance score, and integrating this context with CoT reasoning and few-shot prompting. This approach reduces reliance on prompt quality while efficiently leveraging external knowledge without the need for excessive computational resources, as the integration of external knowledge with CoT prompts does not require LLM fine-tuning.

\section{Methodology}
In this section, we first provide an overview of the \modelname framework. We then describe each component in detail and explain how they integrate to mitigate hallucinations in medical decision making.

\begin{figure*}[ht]
    \centering
    \includegraphics[width=\linewidth]{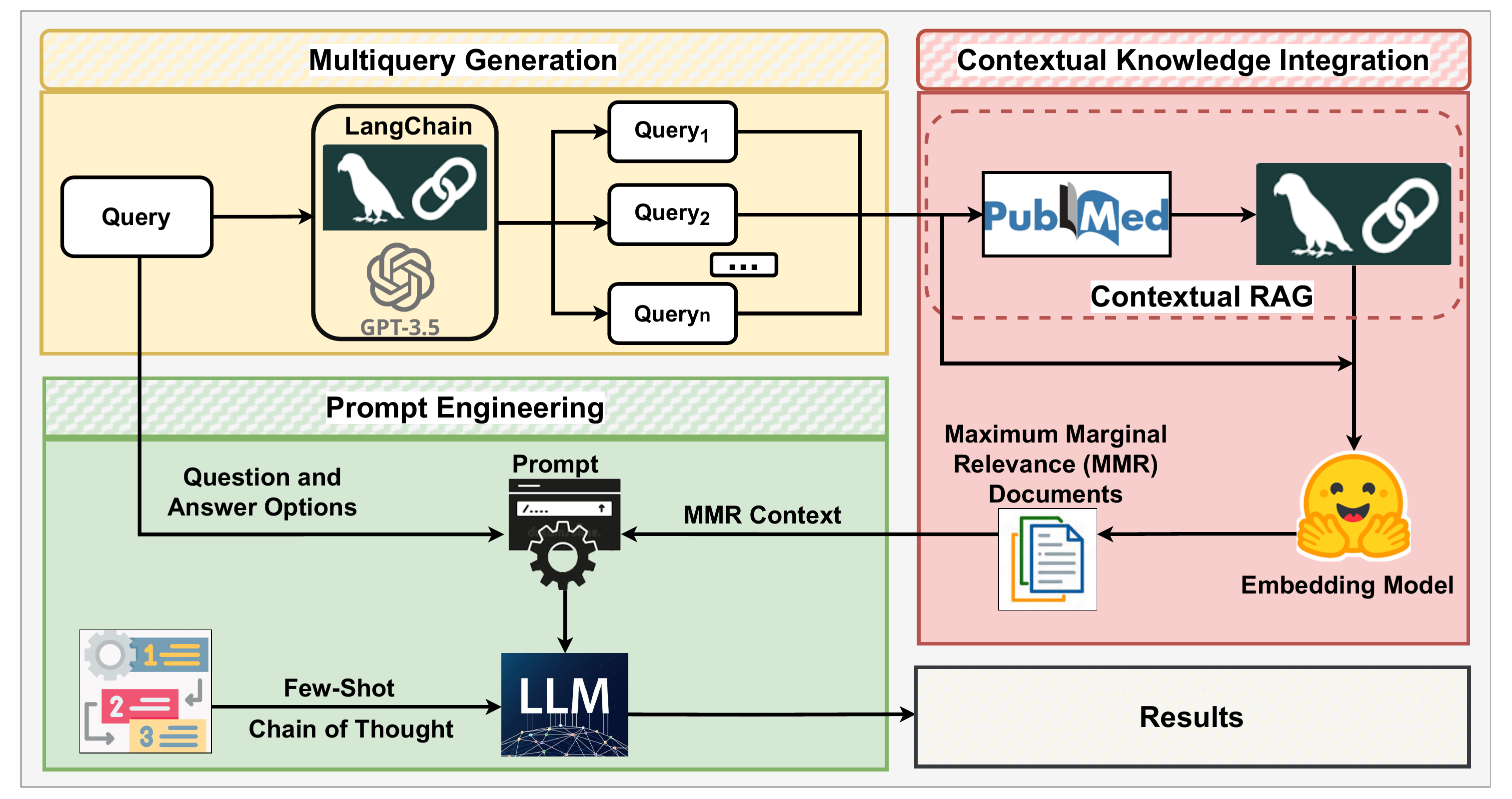}
    \caption{\modelname framework overview. \modelname comprises three key components: multiquery generation, contextual knowledge integration through the maximum marginal relevance-optimized RAG, and few-shot and CoT-based prompt engineering.}
    \label{fig:architecture}
    \vspace{-0.5cm}
\end{figure*}

\subsection{\modelname Framework Overview}

To mitigate hallucination in LLMs during medical decision-making, we introduce \modelname, which integrates many optimization techniques, including query expansion, RAG, MMR-based document selection, few-shot learning, and CoT reasoning. As depicted in Figure~\ref{fig:architecture}, \modelname comprises three core components: multiquery generation, contextual knowledge integration, and prompt engineering.

First, to enhance the contextual understanding~\cite{andriopoulos2023augmenting} of each query, we collect supplementary information from PubMed articles and Wikipedia. Specifically, by generating multiple queries for each question and retrieving information from PubMed, we obtain diverse perspectives and keywords, facilitating a more nuanced understanding of the query.
Second, we leverage LangChain for document retrieval~\cite{langchain}, utilizing embeddings generated by the Instructor-XL model from Hugging Face~\cite{su2022one}. This optimization improves the efficiency of retrieval and comparison, ensuring access to relevant documents. Moreover, we employ the maximum marginal relevance technique to enhance the diversity of the retrieved documents, reducing redundancy and improving the overall quality of information. This ranking process is integrated to refine retrieval results and prioritize the most relevant documents, thereby enhancing the relevance of responses. Additionally, we implement few-shot and CoT prompting to improve the model’s understanding and coherence in responding to the questions, enabling it to generalize more effectively from limited examples and thereby improving performance.

In the following subsections, we will present a detailed breakdown of each component in \modelname and provide an example (see Figure~\ref{fig:example}) to illustrate how \modelname operates.

\subsection{Multiquery Generation}
Multiquery involves generating multiple queries $Q$ for a single query $q$, broadening the scope of perspectives and keywords explored compared to relying solely on a single question and keyword. This approach, exemplified by LangChain, facilitates a more exhaustive exploration of the subject matter, thereby enhancing the effectiveness of information retrieval. Creating multiple queries for a single question becomes crucial as it expands the search parameters, thereby increasing the likelihood of capturing relevant context and information. Given any LLM's challenges in accurately responding to intricate medical queries due to constraints in its training data and domain-specific understanding, generating multiple queries enables the extraction of diverse perspectives and keywords from various sources. This process significantly augments the probability of uncovering pertinent information to effectively tackle the question at hand.

In LangChain, multiquery generation~\cite{langchain} is performed using an LLM, to produce multiple alternative phrasings of a single query, thereby broadening the scope of document retrieval. First, the user integrates the LLM through an API, in our case, ChatGPT-3.5, and LangChain uses its \textit{runnable interface} which is a standardized framework for executing various components, including LLMs to execute the process. A prompt template is predefined to control the structure of the task, ensuring that the generated queries maintain the original query's semantic meaning while altering the wording. After the LLM generates these query variations, a parser called \textit{LineListOutputParser} is used to break down the output into a list of distinct queries, each treated as a separate input for document retrieval. The number and diversity of the queries are customizable by adjusting the prompt or retriever parameters, providing flexibility to generate more focused or diverse query sets. Additionally, the LLM used for this task can easily be changed based on the user's preference, allowing flexibility for different models or integrations.
For instance, consider a medical query regarding a specific drug like Remifentanyl. A singular query may only delve into one aspect, such as the drug's mechanism of action. However, by employing multiple queries, \modelname can explore various facets of the drug, encompassing its pharmacokinetics, side effects, and clinical applications (see the gray block below). Each query targets a distinct aspect, empowering \modelname to retrieve a comprehensive array of pertinent insights from diverse sources, thereby fostering a deeper understanding of the subject matter.

\begin{tcolorbox}[colback=gray!5, colframe=black, boxrule=1pt, rounded corners, title=\textbf{Multiquery Generation Example}, fonttitle=\bfseries]
    \begin{itemize}[leftmargin=0pt]
    \item \textbf{Original Single Query} $q$:
        \begin{itemize}
            \item $q$: What are the characteristics of Remifentanyl?
        \end{itemize}
    \item \textbf{Generated Multiple Queries} $Q$=$\{Q_1, Q_2, ..., Q_n\}$:
        \begin{itemize}
            \item $Q_1$: What are the distinguishing features of Remifentanil?
            \item $Q_2$: How does Remifentanil differ from other opioids, such as Alfentanil?
            \item $Q_n$: What are the unique characteristics of Remifentanil, particularly regarding metabolism, half-life, potency, and dosage adjustments in hepatic and renal disease?
        \end{itemize}
    \end{itemize}
\end{tcolorbox}

\subsection{Contextual Knowledge Integration}
Retrieval augmented generation (RAG) is a novel framework that combines retrieval and generation processes to produce more accurate and contextually relevant responses. It leverages pre-existing knowledge sources, such as databases or documents, to enhance the generation of responses by incorporating retrieved information into the generated text. \modelname adopts RAG to the next level by integrating it with a multiquery approach.
This advanced integration involves employing a multiquery RAG approach, optimizing document retrieval from PubMed by generating diverse queries for each user input. By generating multiple queries, each targeting different aspects of the input question, \modelname can retrieve a broader range of relevant documents from PubMed. By tapping into PubMed's vast repository of peer-reviewed articles and research papers, \modelname enriches its comprehension of medical topics.

RAG is capable of retrieving multiple pieces of information, but the quality and relevance of these results can vary significantly~\cite{yu2024evaluation}. \modelname incorporates a maximum marginal relevance (MMR) scheme to mitigate redundancy and prioritize diversity among the retrieved relevant results $R$. Note that \modelname first expands a single query $q$ into multiple queries $Q$. Instead of using the traditional single-query MMR, We design a multiquery MMR as shown in Equation~\ref{equ:mmr}. By evaluating the relevance of each document $D_i \in R$ based on the new information it contributes compared to previous results, the multiquery MMR effectively reduces duplication while preserving query relevance.

\begin{equation}\label{equ:mmr}
\begin{split}
\text{MMR} \stackrel{\text{def}}{=} \arg \max_{D_i \in R \setminus S} \left[\lambda \cdot \frac{1}{n} \sum_{k=1}^{n} \text{Rel}(D_i, Q_k) \right. \\
\left. - (1 - \lambda) \cdot \max_{D_j \in S} \text{Sim}(D_i, D_j)\right]
\end{split}
\end{equation}

\noindent where $D_i$ is a document in the set of retrieved results $R$ from PubMed; $S$ is the set of already-selected documents; $Q_k$ is the $k$-th query in the generated multiple queries $Q$; $\text{Rel}(D_i, Q_k)$ is the relevance of document $D_i$ to the $k$-th query; $\text{Sim}(D_i, D_j)$ is the similarity between two documents $D_i$ and $D_j$; and $\lambda$ is a parameter that controls the balance between relevance and diversity.

To accurately calculate $\text{Rel}(D_i, Q_k)$ and $\text{Sim}(D_i, D_j)$ in Equation~\ref{equ:mmr}, we generate embeddings for both the retrieved external documents $R$ and the medical queries $Q$, along with their associated answer options. To accomplish this, \modelname utilizes the Hugging Face Instructor-XL model~\cite{su2022one}, an instruction-finetuned text embedding model designed to create task-specific embeddings for diverse applications. This allows \modelname to accurately select and create the most contextually relevant information, retrieved through the multiquery, MMR-optimized RAG process, for the given question and its associated answer choices.

This careful contextual knowledge integration ensures that the responses of \modelname are comprehensive, avoiding unnecessary repetition and leading to higher-quality outputs. The combination of the multiquery RAG approach with MMR optimization significantly improves the efficacy of \modelname retrieval and ranking processes, resulting in reliable responses from our framework.

\subsection{Prompt Engineering}
The \modelname harnesses the capabilities of few-shot learning and CoT reasoning to further refine the context, ensuring both coherence and logical correctness in the responses generated by LLMs. Few-shot prompting enables LLMs to generalize from a limited number of examples, improving their performance even with minimal training data. CoT reasoning, on the other hand, allows LLMs to break down complex medical questions into smaller, manageable components, encouraging step-by-step reasoning that mirrors human thought processes. This approach enhances LLMs' ability to handle nuanced medical queries and provides more accurate and detailed answers. By incorporating few-shot examples and adopting CoT reasoning, \modelname guides LLMs to produce well-structured, contextually relevant, and logically sound responses.

\subsection{An Example of How \modelname Works}

\begin{figure*}[ht]
    \centering
    \includegraphics[width=\linewidth]{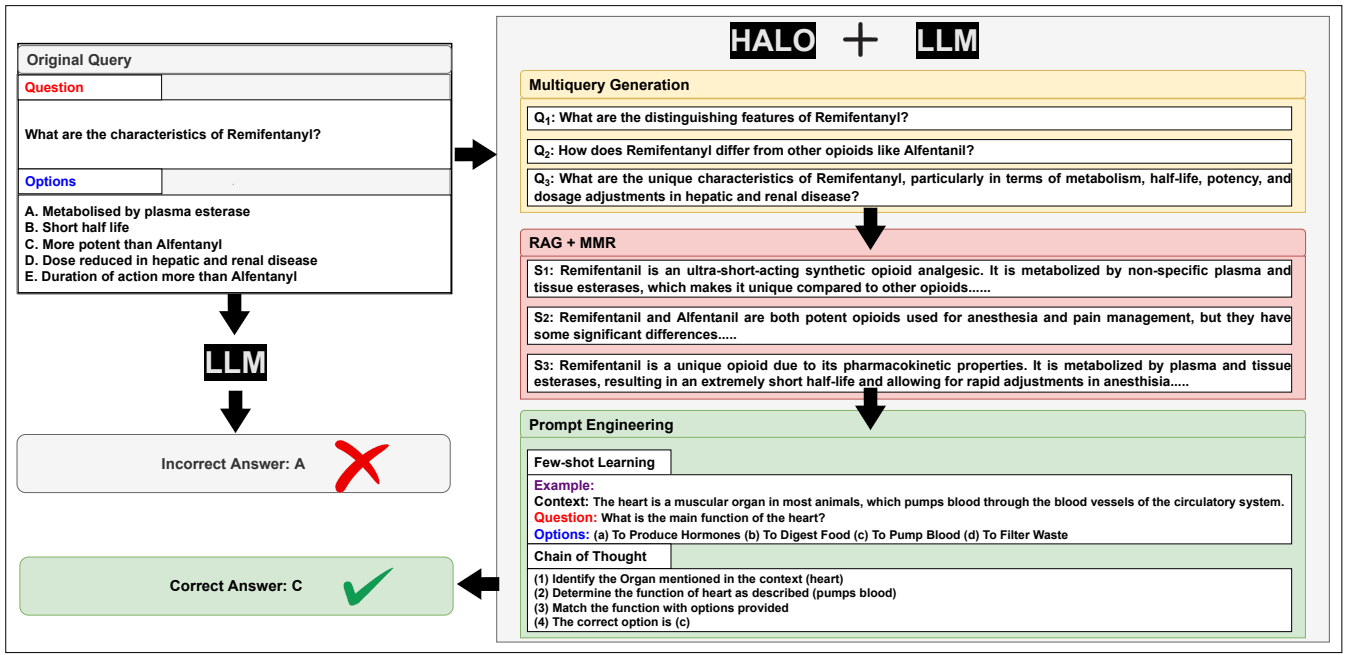}
    \caption{Example of \modelname framework applied to a medical question from the MedMCQA dataset}
    \label{fig:example}
\end{figure*}

Figure~\ref{fig:example} illustrates the \modelname framework applied to enhance the accuracy of medical question answering for a sample query. In this example, a question regarding the characteristics of Remifentanyl is posed, and the baseline model incorrectly selects the answer ``A" instead of the correct answer ``C". Using our framework, three alternative queries (highlighted in yellow) are generated through the LangChain Multiquery Retriever to expand the search scope. Relevant information is then retrieved from PubMed (highlighted in red), enriching the context and providing diverse perspectives. The framework further incorporates few-shot learning and CoT reasoning by offering structured prompts and examples (highlighted in green) to guide the model towards more accurate responses.

\section{Evaluation}
This section first introduces commonly used evaluation metrics for assessing hallucination in LLMs and provides a detailed explanation for the specific metric chosen for this work. Next, we describe the dataset and the LLMs used in our testing framework. We also present the experimental results of \modelname on the dataset, comparing its performance against several commercial and open-source LLMs. Finally, we select neurological disorders as a case study to demonstrate the effectiveness of \modelname.

\subsection{Evaluation Metrics}
The evaluation of hallucination detection can be approached in various ways, depending on the specific task, including fact-based metrics, classifier-based metrics, QA-based metrics, uncertainty estimation, and prompting-based metrics~\cite{huang2023survey}. \textbf{Fact-based metrics}~\cite{thorne2018fever}, checks if the facts in the model's responses match the facts in the original content. It is like comparing two lists to see how many items they have in common. While this technique is useful, it often misses the nuances and context-specific details with poor correlation. \textbf{Classifier-based metrics}~\cite{jeong2024olaph} involve using natural language inference models that can determine if the model's response logically follows from the original content. Consequently, classifier-based metrics require extensive training data to be effective. \textbf{QA-based metrics}~\cite{sharma2024retrieval} involves asking questions based on the original content and comparing the model's answers to the correct ones. \textbf{Uncertainty estimation} technique measures how confident the model is in its answers, this type of detection is used in zero-resource settings~\cite{huang2023survey}. If the model is unsure, there is a higher chance it might be making things up. This method can be useful but often needs to be combined with other metrics for comprehensive evaluation. \textbf{Prompting-based metrics}~\cite{yu2022legal,zhu2023promptcblue} uses specific prompts to guide the model in evaluating its own responses for accuracy. The model is given examples or specific instructions to help it check if its answers are faithful to the source content. This method can enhance the accuracy of responses but depends heavily on the quality of the prompts.

\modelname is evaluated using QA-based metrics, as they directly measure the model's understanding and ability to generate accurate responses. This approach was also chosen for our evaluation due to its direct applicability to the MedMCQA dataset. To guide the model in providing consistent and accurate responses, we used a few-shot prompting strategy, giving examples of the correct answer format. In addition, we cleared the chat history after each question to prevent previous interactions from influencing subsequent answers. This approach allowed us to identify hallucinations by ensuring the model's responses were directly comparable to the expected answers.

\subsection{MedMCQA Dataset}
The MedMCQA dataset is specifically designed to address real-world medical entrance exam questions, making it highly relevant for training and evaluating QA models in the medical domain. This ensures that the dataset reflects the types of questions and challenges that medical practitioners may encounter in their professional practice.

\begin{figure}
    \centering
    \includegraphics[width=0.6\linewidth]{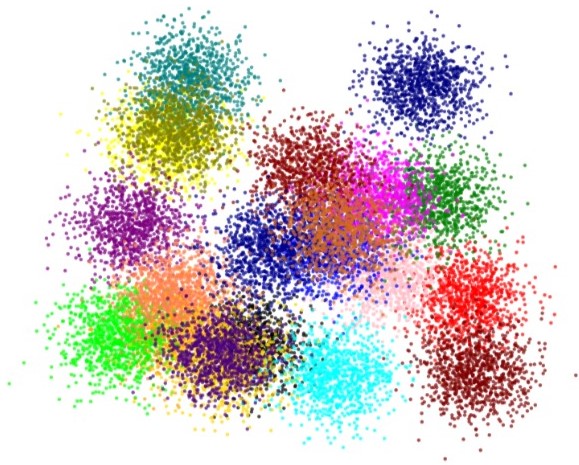}
    \caption{Distribution of MedMCQA questions across 21 medical subjects, covering a wide range of topics}
    \label{fig:subjectvariations}
\end{figure}

The MedMCQA dataset consists of multiple-choice questions (MCQs), which are a common format in medical exams~\cite{pal2022medmcqa, umapathi2023med}. This format requires models to not only identify the correct answer but also reason and differentiate it from the other distractors, adding an additional layer of complexity to the task. The large size of the dataset, with over 194,000 MCQs, provides a rich and diverse set of examples for training and evaluation. The dataset is divided with approximately 53\% for DEV and 47\% for TEST categories, with medical subjects and topics evenly distributed across both sets for balanced evaluation. This helps in capturing a wide range of medical knowledge, scenarios, and variations in question styles, ensuring that models are exposed to a comprehensive set of challenges. The questions in the MedMCQA dataset are created by human experts, ensuring their quality and reliability. This expert curation guarantees that the dataset represents accurate and credible medical knowledge, enhancing the validity of the evaluation and the trustworthiness of the results. Figure~\ref{fig:subjectvariations} presents the t-SNE visualization~\cite{van2008visualizing} of MedMCQA's 21 medical subjects, demonstrating comprehensive coverage across a wide range of knowledge areas, encompassing 2,400 internal medical topics. The distribution of subjects in the dataset indicates a fair allocation of focus across different areas of medical expertise.

\subsection{LLM Models}
We evaluated the performance of \modelname on both commercial and open-source LLMs, including ChatGPT-3.5-16K, Llama-3.1 8B, and Mistral 7B.

\subsubsection{ChatGPT-3.5-16K}
We have chosen to use the ChatGPT-3.5-16K model in our evaluation for several reasons. First, this model has a maximum context length of 16,384 tokens~\cite{openai_gpt35_turbo}, which allows it to process and understand larger inputs and provide more detailed outputs, making it well-suited for handling the intricacies of medical knowledge and providing contextually relevant responses~\cite{rosol2023evaluation}. The ChatGPT-3.5-16K model has also been specifically fine-tuned for instruction-based tasks~\cite{openai_gpt35_turbo,anakin_gpt35_turbo}, making it well-suited for our objective of enhancing medical question answering. By utilizing the ChatGPT-3.5-16K model, we aim to leverage its advanced capabilities and specialized training in providing comprehensive information to medical professionals and users seeking medical insights.

\subsubsection{Llama-3.1 8B}
Llama-3.1 8B is an open-source powerful and adaptable NLP model, optimized for a wide range of tasks with high efficiency and accuracy. As one of the most recently released Llama models, which has been optimized for a wide variety of instructions, it excels at following instructions precisely~\cite{raza_Llama_3_1}. As an open-source model, it offers transparency and flexibility, allowing us to customize and optimize it for our specific needs.

\subsubsection{Mistral 7B}
Mistral 7B~\cite{jiang2023mistral} is an efficient, open-source language model that excels in complex NLP tasks with a smaller parameter size, making it lightweight and fast. It offers a good balance between computational efficiency and performance, especially in resource-constrained environments. After evaluating ChatGPT-3.5-16k and Llama-3.1 8B for our medical QA task, we selected Mistral 7B to meet the requirement for a model capable of delivering robust performance even in environments with limited resources.

\subsection{Experimental Results on 21 Medical Subjects}

Using the QA-based evaluation metric for hallucination detection, which aligns with Faithfulness Hallucination Detection as detailed in~\cite{huang2023survey,ji2023survey}, we first evaluated the performance of BERT-based~\cite{devlin2018bert} and Codex-based~\cite{chen2021evaluating} language models on the MedMCQA dataset. As the dataset is split by the authors into two subsets, DEV (53\%) and TEST (47\%), we report the accuracies for both. As shown in Table~\ref{tab:results}, BERT Base, BioBERT, SciBERT, and PubMedBERT exhibit very low accuracies on both the DEV and TEST subsets~\cite{pal2022medmcqa}. In contrast, Codex 5-shot CoT and MQ-SequenceBERT~\cite{lievin2023variational} demonstrate improvements due to advanced techniques~\cite{umapathi2023med,ponce2024large}, such as few-shot learning and sequence-based enhancements.

\begin{table}[htbp]
    \centering
    \begin{tabular}{lcc}
        \toprule
        \textbf{Models} & \textbf{DEV Acc.} & \textbf{TEST Acc.} \\
        \hline
        BERT Base & 0.35 & 0.33 \\
        BioBERT & 0.38 & 0.37 \\
        SciBERT & 0.39 & 0.39 \\
        PubMedBERT & 0.40 & 0.41 \\
        Codex 5-shot CoT & 0.63 & 0.60 \\
        MQ-SequenceBERT & 0.68 & 0.60 \\
        ChatGPT-3.5 & 0.56 & 0.44 \\
        Llama-3.1 8B & 0.44 & 0.38 \\
        Mistral 7B & 0.37 & 0.28 \\
        \textbf{\modelname + ChatGPT 3.5} & \textbf{0.70} & \textbf{0.65} \\
        \textbf{\modelname + Llama-3.1 8B} & \textbf{0.66} & \textbf{0.60} \\
        \textbf{\modelname + Mistral 7B} & \textbf{0.58} & \textbf{0.55} \\
        \bottomrule
    \end{tabular}
    \caption{Performance of LLMs on the MedMCQA DEV (53\%) and MedMCQA TEST (47\%) subsets}\label{tab:results}
\end{table}

One of the most widely used LLMs, ChatGPT-3.5-turbo-16K, achieved only 56\% accuracy—significantly lower than the typical 90\% or higher accuracy achieved by human medical professionals~\cite{umapathi2023med}. After applying \modelname, accuracy improved from 44\% to 65\% on the TEST subset (and from 56\% to 70\% on the DEV subset). Similarly, the accuracy of the Llama-3.1 8B model improved from 38\% to 60\% with the application of \modelname on the TEST subset (and from 44\% to 66\% on DEV subset). The Mistral 7B model also showed a boost in accuracy, increasing from 28\% to 55\% on the TEST subset (and 37\% to 58\% on DEV subset), though it still fell short of expectations. Notably, our \modelname framework achieved the highest accuracies—70\% on the DEV set and 65\% on the TEST set—demonstrating its superior performance and effectiveness in addressing medical QA tasks.

\begin{figure*}[ht]
    \centering
    \includegraphics[width=\linewidth]{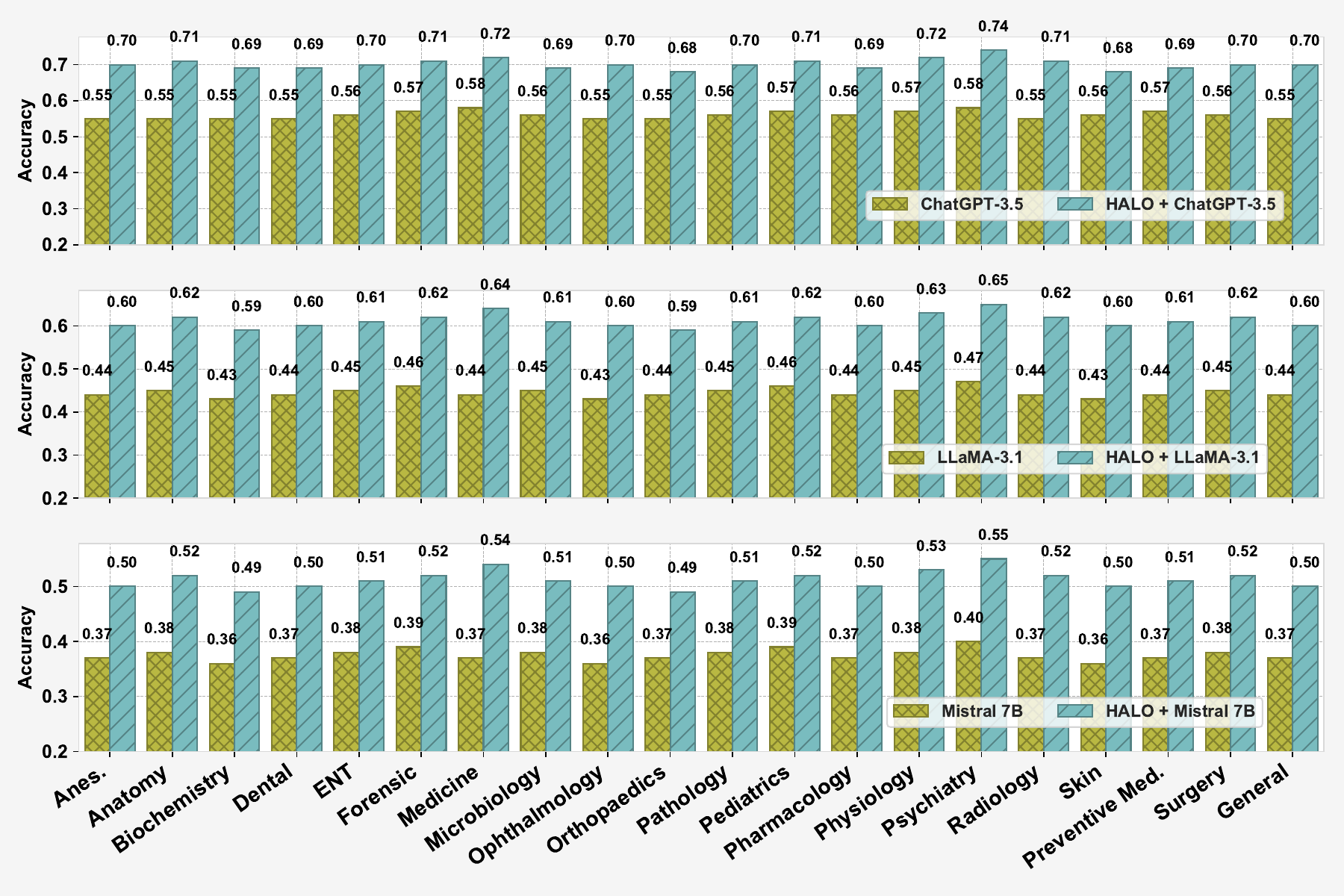}
    \caption{Comparative analysis of the LLMs (ChatGPT-3.5, Llama-3.1 8B, Mistral 7B) and the \modelname framework on LLMs accuracies across 21 subjects in MedMCQA dataset}
    \label{fig:performance}
    \vspace{-0.4cm}
\end{figure*}

To evaluate the generalizability of \modelname across different medical subjects, we evaluated its performance on all 21 subjects (e.g., anatomy and dental) covered by the MedMCQA dataset. Figure~\ref{fig:performance} presents the comparative accuracy of LLMs and \modelname across these subjects. ChatGPT-3.5, Llama-3.1 8B, and Mistral 7B also showed consistent and steady improvements in performance across all subjects. These results highlight the effectiveness of \modelname in generating and prioritizing relevant contextual information, further demonstrating its robustness and superiority over strong baselines on the MedMCQA dataset.

\subsection{Case Study on Neurological Disorders}
Neurological disorders present diagnostic and treatment challenges due to their complex and variable symptoms, necessitating broad medical knowledge for informed decision-making. We began by filtering relevant questions from the MedMCQA dataset, which is organized by subjects like Medicine, Psychiatry, General, and Preventive Medicine using keywords such as ``aphasia'', ``dementia'', and ``mild cognitive impairment''. Then we applied \modelname on these questions. The sample question with the multiquery generation is shown in the following gray box, where multiple queries are generated to explore different aspects of the aphasia diagnosis, enhancing the model’s ability to understand and retrieve relevant information.

\begin{tcolorbox}[colback=gray!10, boxrule=1pt, colframe=black, title=Example Question and Multiquery Generation]
\textbf{Aphasia Diagnosis Question $q$: } \\
A man comes with aphasia. He is unable to name things and repetition is poor. However, comprehension, fluency, and understanding of written words are unaffected. He is probably suffering from:

A. Anomic aphasia

B. Broca's aphasia

C. Transcortical sensory aphasia

D. Conduction aphasia

E. Global aphasia

\tcblower

\textbf{Multiquery Generation}

\textbf{$Q_1$}: What are the common types of aphasia, and how do their symptoms differ? 

\textbf{$Q_2$}: What type of aphasia is characterized by the inability to name things and poor repetition, but with unaffected comprehension, fluency, and understanding of written words?

\textbf{$Q_3$}: How do different aphasia types impact a patient's ability to communicate, and what are the key signs to look for?
\end{tcolorbox}

\begin{tcolorbox}[colback=gray!10, boxrule=1pt, colframe=black, title=Chain of Thought (CoT) Reasoning]
\textbf{1. Identify the main subject or keyword in the question:} 
The subject is a patient with aphasia exhibiting poor repetition and naming ability but preserved fluency and comprehension.

\textbf{2. Understand the relationship between the subject and the question asked:} 
The goal is to identify which type of aphasia corresponds to these specific symptoms.

\textbf{3. Extract relevant information from the context:}

\vspace{0.1cm}
\includegraphics[width=0.10\textwidth]{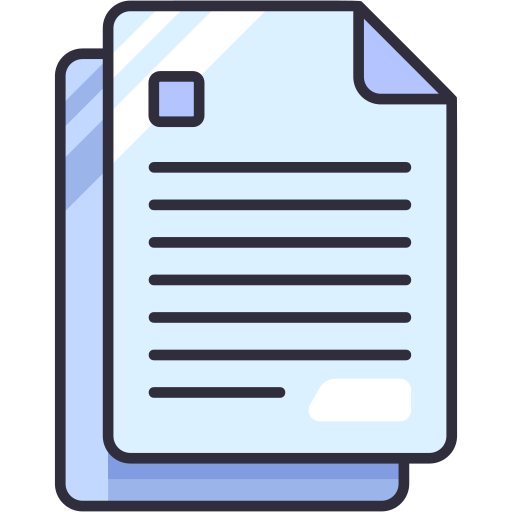}
\includegraphics[width=0.10\textwidth]{figures/doc.png} 
\includegraphics[width=0.10\textwidth]{figures/doc.png} 
\vspace{0.1cm}

Contextual documents/information retrieved.

\textbf{4. Analyze each option:}

\textbf{A. Anomic aphasia:} Naming difficulty fits, but repetition is unaffected, so this is incorrect.

% \textbf{B. Broca’s aphasia:} Non-fluent speech contradicts the patient’s fluent speech, so this is ruled out.

% \textbf{C. Transcortical sensory aphasia:} Comprehension is impaired in this condition, which does not match the patient.

\textbf{B. C. D. E. }...

% \textbf{D. Conduction aphasia:} Matches all the symptoms poor repetition, difficulty naming objects, but fluent speech and intact comprehension.

% \textbf{E. Global aphasia:} Does not fit because the patient retains fluency and comprehension.

\textbf{5. Provide the answer:} The correct answer is ``D'' Conduction aphasia, as it aligns perfectly with the described symptoms.
\end{tcolorbox}

The CoT reasoning of the aphasia diagnosis example question is outlined in the above box, detailing the logical steps to reach the correct answer taken by the LLM. The CoT process begins with identifying the main subject (a patient with aphasia) and understanding the objective by determining the type of aphasia. Next, each option is analyzed based on the described symptoms. The MMR content extracted from PubMed is reviewed by the ChatGPT-3.5 to compare the characteristics of different aphasia types. After this analysis, it is determined that option ``D. Conduction aphasia'', is the correct choice, as it aligns best with the patient's symptoms of poor repetition, difficulty with naming, but preserved fluency and comprehension. Finally, ChatGPT-3.5 generates the output by adhering to these CoT steps and utilizing the content.

\begin{table}[htbp]
\centering
\begin{tabular}{lcc}
\toprule
\textbf{Models} & \textbf{w/o \modelname Acc.} & \textbf{w/ \modelname Acc.} \\
\midrule
ChatGPT-3.5-16K & 0.49 & \textbf{0.65} \\
Llama-3.1 8B & 0.40 & \textbf{0.58} \\
Mistral 7B & 0.34 & \textbf{0.48} \\
\bottomrule
\end{tabular}
\caption{The performance of \modelname for neurological disorder related questions from MedMCQA dataset}\label{tab:results_neuro}
\vspace{-0.2cm}
\end{table}

\subsubsection{Case Study Results}
We performed an ablation evaluation on ChatGPT 3.5, Llama-3.1 8B, and Mistral 7B to establish baseline accuracies, which were 49\%, 40\%, and 34\%, respectively. Applying \modelname improved these accuracies to 65\%, 58\%, and 48\% as shown in Table~\ref{tab:results_neuro}. This case study demonstrates that \modelname effectively boosts the performance of mitigating hallucination in neurological disorder related medical decisions.

\vspace{-0.3cm}
\section{Conclusion}
This paper introduces \modelname, a novel open-source framework designed to identify and mitigate hallucinations in LLMs, with a specific emphasis on the medical domain. By incorporating advanced techniques such as multiquery generation, contextual knowledge integration, and prompt engineering, \modelname demonstrates significant improvements in mitigating hallucinations. To thoroughly evaluate the framework, we tested \modelname across multiple LLMs, including ChatGPT-3.5, Llama-3.1 8B, and Mistral 7B, as well as six other BERT- and Codex-based models. These evaluations on the MedMCQA dataset highlighted notable enhancements in both accuracy and reliability when \modelname was integrated, achieving consistent improvements across all 21 medical subjects. In addition, a case study on neurological disorders further underscored \modelname's effectiveness in addressing specific disease-related queries.

\bibliography{aaai25}

\end{document}